# Structured variable selection in support vector machines

**Seongho Wu and Hui Zou**[*]

*School of Statistics
University of Minnesota
Minneapolis, MN 55455
e-mail:* `swu@stat.umn.edu`*;* `hzou@stat.umn.edu`

**Ming Yuan**

*School of Industrial and Systems Engineering
Georgia Institute of Technology
Atlanta, GA 30332-0205
e-mail:* `myuan@isye.gatech.edu`

**Abstract:** When applying the support vector machine (SVM) to high-dimensional classification problems, we often impose a sparse structure in the SVM to eliminate the influences of the irrelevant predictors. The lasso and other variable selection techniques have been successfully used in the SVM to perform automatic variable selection. In some problems, there is a natural hierarchical structure among the variables. Thus, in order to have an interpretable SVM classifier, it is important to respect the heredity principle when enforcing the sparsity in the SVM. Many variable selection methods, however, do not respect the heredity principle. In this paper we enforce both sparsity and the heredity principle in the SVM by using the so-called structured variable selection (SVS) framework originally proposed in [20]. We minimize the empirical hinge loss under a set of linear inequality constraints and a lasso-type penalty. The solution always obeys the desired heredity principle and enjoys sparsity. The new SVM classifier can be efficiently fitted, because the optimization problem is a linear program. Another contribution of this work is to present a nonparametric extension of the SVS framework, and we propose nonparametric heredity SVMs. Simulated and real data are used to illustrate the merits of the proposed method.

Received September 2007.

## 1. Introduction

The support vector machine (SVM) is a widely used classification method. Let $x$ denote a generic feature vector. The class labels, $y$, are coded as $\{1, -1\}$. For a given training data set $\{x_i, y_i\}, i = 1, 2, \ldots, n$, the SVM can be expressed in a penalized hinge loss formulation (cf. [11] and [15])

$$(\hat{\beta}, \hat{\beta}_0) = \arg\min_{\beta,\beta_0} \sum_{i=1}^{n} \left[1 - y_i(x_i^T \beta + \beta_0)\right]_+ + \lambda \|\beta\|_2^2, \quad (1.1)$$

where the subscript "+" means the positive part ($z_+ = \max(z, 0)$). The SVM classifier is $\text{Sign}(\hat{\beta}_0 + x^T \hat{\beta})$. It is now well known that by imposing some structure in the SVM, we could significantly enhance its classification performance and obtain a more interpretable model [11]. For example, when the dimension of the predictors is high and there are many irrelevant predictors, imposing sparsity in $\beta$ via an automatic variable selection procedure can significantly enhance classification performance of the SVM. Various variable selection proposals have been introduced in recent years to encourage sparsity in $\beta$ for the SVM. See [25] and references therein. In particular, Bradley and Mangasarian [1] and Zhu et al. [24] suggested to replace the quadratic penalty in (1.1) with the lasso (or $l_1$) penalty:

$$\sum_{i=1}^{n} \left[1 - y_i(x_i^T \beta + \beta_0)\right]_+ + \lambda \|\beta\|_1. \quad (1.2)$$

---

[*]Corresponding author. Zou's research is supported by National Science Foundation grant DMS 0706733.

Similar to the lasso [13] for linear regression, the lasso penalty encourages some of the $\beta$ coefficients to exact zero and therefore perform variable selection.

Despite their successes, these general-purpose variable selection methods do not take advantage of the possible interrelationship among features. Consider for example a quadratic classifier with explanatory variables $z_1, z_2, \ldots, z_q$:

$$\beta_1 z_1 + \ldots + \beta_q z_q + \beta_{11} z_1^2 + \beta_{12} z_1 z_2 + \ldots + \beta_{q,q-1} z_q z_{q-1} + \beta_{qq} z_q^2. \quad (1.3)$$

In employing the $l_1$ SVM to learn the $\beta$ coefficients, one may consider using $x \equiv (z_1, \ldots, z_q, z_1 z_2, \ldots, z_{q-1} z_q, z_1^2, \ldots, z_q^2)$ as the derived variables in (1.2). In doing so, we neglect the difference between quadratic effects and linear effects. In situations like this, it is desirable to invoke the effect heredity principle [18]. There are two popular versions of the effect heredity [3]. Under the *strong heredity*, for a two-factor interaction effect $z_i z_j$ to be active both its parent effects, $z_i$ and $z_j$, should be active, whereas under the *weak heredity* only one of its parent effects needs to be active. Likewise, one may also require that $z_j^2$ is allowed to be active only if $z_j$ is active. In this paper we develop a new method that can simultaneously impose the sparse structure and the heredity structure in the SVM model.

Earlier interests in the heredity principle came from the analysis of designed experiments where heredity principle had proven to be powerful tools in resolving complex aliasing patterns (cf. [3], [4] and [9]). The heredity principle was routinely followed in general regression problems as well. Efron et al. [7] and Turlach [14] discussed how to enforce the strong heredity principle in the efficient Lars algorithm. Later, Yuan, Joseph and Lin [19] proposed more flexible ways of incorporating the strong and weak heredity principles in linear regression. Zhao,

Rocha and Yu [23] presented the *Composite Absolute Penalties* which can produce a hierarchical model. Choi and Zhu [5] proposed a penalization method for enforcing the strong heredity principle in fitting a regression model. However, these earlier methods are primarily designed for the linear regression model. It is not clear how to generalize them to handle other models such as the SVM considered in the present paper and still retain their computational efficiency.

More recently, Yuan, Joseph and Zou [20] formalized the concept of *structure variable selection* to describe general hierarchical structures among variables with traditional heredity principles as special cases when doing variable selection. They argue that appropriately accounting for the general hierarchical structure among variables not only enhances the model interpretability but also leads to improved estimation and prediction. The SVS framework gives a unified treatment of the linear regression model and generalized linear models. In addition, the SVS framework permits a very efficient implementation and enjoys nice theoretical properties.

In this paper, we propose to adopt the SVS framework to simultaneously incorporate the heredity principle and sparsity into the support vector machine in a way that retains the computational advantages of the SVM. The main idea is to introduce a scaling parameter to each effect and then enforce the hierarchical relationships among predictors and sparsity by a set of linear inequality constraints on the corresponding scaling parameters. As a result, the optimization problem is a linear program and can be very efficiently solved using standard linear programming techniques. Our approach can handle both strong and weak heredity principles. Furthermore, we propose a nonparametric extension of the SVS framework based on which we develop nonparametric heredity SVMs.

The rest of the paper is organized as follows. In the next section, we describe how to employ the SVS idea to incorporate the strong heredity principle into the SVM. The weak heredity principle can be implemented in a similar fashion with an additional *convex relaxation step*, in order to preserve the computational efficiency. In Section 3 we propose the nonparametric heredity SVMs. Section 4 contains some discussion.

## 2. The Generalized Garrote and Heredity Principles

### 2.1. Method

Breiman's nonnegative garrote [2] is perhaps the first method in the literature that uses an $l_1$ constraint to perform variable selection in linear regression models. As an extension of the original nonnegative garrote, the generalized garrote is first introduced in [20] to build the SVS framework. Here we show that the generalized garrote idea can be used in the support vector machine as well. To provide the readers a complete picture, we first introduce the basic idea of the garrote in the context of the SVM. Suppose we have computed the $l_2$ SVM coefficients $\hat{\beta}$, then we introduce a scaling parameter $\theta_j$ for each predictor $x_j$ and then solve the following optimization problem

$$\min_{\{\theta_j\},\beta_0} \sum_{i=1}^n \left[1 - y_i(\sum_{j=1}^p x_{ij}\hat{\beta}_j\theta_j + \beta_0)\right]_+ \quad (2.1)$$
$$\text{subject to} \quad \sum_{j=1}^p \theta_j \leq M$$
$$\theta_j \geq 0 \quad \forall j,$$

where $M$ is the garrote shrinkage parameter. The new classifier is $\text{Sign}(\hat{\beta}_0 + \sum_{j=1}^p x_j\hat{\beta}_j\hat{\theta}_j)$. To compare it with the $l_1$ SVM, we consider another equivalent formulation of (2.1)

$$\min_{\{\theta_j\},\beta_0} \sum_{i=1}^n \left[1 - y_i(\sum_{j=1}^p x_{ij}\hat{\beta}_j\theta_j + \beta_0)\right]_+ + \lambda \sum_{j=1}^p \theta_j \quad (2.2)$$
$$\text{subject to} \quad \theta_j \geq 0 \quad \forall j.$$

When $M$ or $\lambda$ is properly chosen, some $\hat{\theta}_j$s will be shrunk to zero, and thus the corresponding predictors ($x_j$s) will be deleted from the classifier. Therefore, the garrote performs variable selection in a way similar to the lasso.

The garrote received little attention in the literature compared to the enormous popularity of the lasso. Recently, Yuan and Lin [22] showed the garrote enjoys excellent finite sample performance if we use some regularized estimators as the initial estimator. The biggest advantage of the garrote, however, is its flexibility. We can easily modify the garrote by adding other linear constraints on the scaling parameters to meet some special requirements, such as the heredity principle.

We adopt some notation from [20] to formally describe general hierarchical structures among variables. Suppose the dimension of the predictor set is $p$. The hierarchical relationships among predictors can be represented by sets $\{\mathcal{D}_j : j = 1,\ldots,p\}$, where $\mathcal{D}_j$ contains the parent effects of the $j$th predictor. Consider, for example, the predictors in model (1.3). The $q+1$th predictor is $x_{q+1} = z_1 z_2$ and its parent effects are $x_1 = z_1$ and $x_2 = z_2$. Thus $\mathcal{D}_{q+1} = \{1,2\}$.

The strong heredity principle says that if the $j$th predictor can be considered for inclusion, all elements of $\mathcal{D}_j$ must be included. Note that in the garrote model, the $j$th predictor is included if and only if its scaling parameter is nonzero. To further incorporate the strong heredity principle, we generalize the garrote as follows

$$\min_{\{\theta_j\},\beta_0} \sum_{i=1}^n \left[1 - y_i(\sum_{j=1}^p x_{ij}\hat{\beta}_j\theta_j + \beta_0)\right]_+ + \lambda \sum_{j=1}^p \theta_j \quad (2.3)$$
$$\text{subject to} \quad \theta_j \geq 0 \quad \forall j$$
$$\text{and} \quad \theta_j \leq \theta_r, \quad \forall r \in \mathcal{D}_j, \ \forall j. \quad (2.4)$$

We have imposed a set of inequality constraints on the scaling parameters, besides the $l_1$ constraint which ensures the sparsity of the estimates. Note that if $\theta_j > 0$, these linear inequalities in (2.4) force the scaling parameters in $\mathcal{D}_j$ to be positive. Therefore, the resulting model always obeys the strong heredity principle. Furthermore, all the constraints are linear in terms of the scaling parameters, and the feasible region under these constraints is convex. Therefore, solving (2.3) remains a linear program.

The same idea can be applied to impose the weak heredity principle. The weak heredity principle says that if the $j$th variable is included in the model, at least one of the elements of $\mathcal{D}_j$ must be included. Observe that

$$\max_{r \in \mathcal{D}_j} \theta_r > 0 \Leftrightarrow \text{at least one } \theta_r > 0, r \in \mathcal{D}_j \text{ and } \forall j.$$

We could consider the following optimization problem

$$\min_{\{\theta_j\},\beta_0} \sum_{i=1}^n \left[1 - y_i(\sum_{j=1}^p x_{ij}\hat{\beta}_j\theta_j + \beta_0)\right]_+ + \lambda \sum_{j=1}^p \theta_j \quad (2.5)$$

subject to $\theta_j \geq 0 \quad \forall j$

and $\theta_j \leq \max_{r \in \mathcal{D}_j} \theta_r, \quad \forall j.$ (2.6)

It is easy to see that the solution always obeys the weak heredity principle. However, the feasible region under such constraints is no longer convex. It is well known that non-convexity may cause various computational problems such as local minimizer and instability of the solution, etc. To overcome the non-convexity issue, we suggest to use the convex envelop of these constraints for the weak heredity principle

$$\min_{\{\theta_j\},\beta_0} \sum_{i=1}^n \left[1 - y_i(\sum_{j=1}^p x_{ij}\hat{\beta}_j\theta_j + \beta_0)\right]_+ + \lambda \sum_{j=1}^p \theta_j \quad (2.7)$$

subject to $\theta_j \geq 0 \quad \forall j$

and $\theta_j \leq \sum_{r \in \mathcal{D}_j} \theta_r, \forall j.$ (2.8)

Note that under (2.8) $\theta_j > 0$ implies that $\sum_{r \in \mathcal{D}_j} \theta_r > 0$ and therefore at least one of its parents needs to be included in the model, which implies that the resulting model obeys the weak heredity principle. Since the constraints in (2.8) are linear and the feasible region under the constraints in (2.7) is convex, solving (2.7) remains a linear program.

For the purpose of presentation, we refer the new SVMs defined in (2.3) and (2.7) to as SHSVM and WHSVM, respectively.

### 2.2. Numerical studies

We use numerical examples to demonstrate the benefits of incorporating heredity principles into the SVM model.

In each simulated example, we generated 100 datasets, each with training samples of sizes $n = 50, 100$, and $200$, and an independent test sample of size 10000. In a benchmark example, 100 random partitions of the original data were created, each with a training sample and a test sample. In each example, all classifiers were fitted on a training sample and their generalization errors were computed on a test sample. Here the generalization error of a classifier $f$ is $\Pr(yf(x) < 0)$ under 0–1 loss. The Bayes rule minimizes the generalization error and its error is called the Bayes error (risk). Note that the Bayes rule is $\arg\max_{c\in\{1,-1\}} \Pr(y = c|x)$ which is unknown in practice. In our simulation study we can compute the Bayes error because we know the true model. We reported the Bayes error and the averaged smallest generalization error of each competitor, thus avoiding the extra level of complexity in the comparison caused by the tuning parameter selection.

We first consider three simulation models. In the first example the true model obeys the strong heredity principle, while in the second example the true model obeys the weak heredity principle. The third example concerns the situation when the true model does not obey any heredity principle.

**Simulation example 1.** In the first set of simulation, the generated explanatory variables $z_1, \ldots, z_7$ are standard normal, where the correlation between $z_r$ and $z_j$ is $\rho^{|r-j|}$, $\rho = 0, 0.5$. The class labels are generated from a logistic regression model

$$\log\left(\frac{\Pr(y = 1|z_1, \ldots, z_7)}{\Pr(y = -1|z_1, \ldots, z_7)}\right) = 2z_1 + 4z_3 + 3z_1z_3 + 1.$$

The predictor set for fitting the SVMs is $\{z_j, z_r z_j, z_j^2\}$, $r, j = 1, \ldots, 7$. The predictor $z_r z_j$ represents the interaction between predictors $z_r$ and $z_j$, thus its parent effects are $z_r$ and $z_j$. The predictor $z_j^2$ represents the quadratic effect of $z_j$. We include the quadratic effect only if the linear main effect is included. Let $\theta_j$ and $\theta_{jj}$ be the scaling parameters for $z_j$ and $z_j^2$, respectively. Let $\theta_{rj}$ be the scaling parameter for $z_r z_j$ ($r \neq j$). Then the linear constraints in (2.4) become

$$\theta_{rj} \leq \theta_r \quad \text{and} \quad \theta_{rj} \leq \theta_j, \quad \forall r \neq j, \quad r, j = 1, \ldots, 7$$
$$\theta_{jj} \leq \theta_j \qquad j = 1, \ldots, 7.$$

The simulation results are summarized in Table 1. From Table 1 we see that the SHSVM significantly outperforms the $l_1$ and $l_2$ SVMs in terms of classification accuracy regardless of sample sizes, although the differences get smaller as sample sizes increase. We also computed the frequency that the fitted $l_1$ SVM obeys the strong heredity principle, as reported in the last column on Table 1. The low frequency indicates that the $l_1$ SVM is not appropriate when a strong heredity model is in demand.

**Simulation example 2.** In the second set of simulation, we use the same setup in example 1, except that the class labels are generated from a logistic regression model

$$\log\left(\frac{\Pr(y = 1|z_1, \ldots, z_7)}{\Pr(y = -1|z_1, \ldots, z_7)}\right)$$
$$= 3.5z_1 + 3z_1z_2 + 2.5z_1z_3 + 2z_1z_4 + 1.5z_1z_5 + z_1z_6 + 1.$$

This model obeys the weak heredity principle and violates the strong heredity principle. In order to fit the WHSVM, we note that the linear constraints in (2.8) become

$$\theta_{rj} \leq \theta_r + \theta_j \qquad \forall r \neq j, \quad r, j = 1, \ldots, 7$$
$$\theta_{jj} \leq \theta_j \qquad j = 1, \ldots, 7.$$

TABLE 1
*Simulation example 1: The true model obeys the strong heredity principle. Compare the classification accuracy of the SHSVM, $l_2$ SVM and $l_1$ SVM. The numbers in parentheses are standard errors and the frequency is the number of times the fitted $l_1$ SVM obeys the strong heredity principle in 100 replications.*

| $\rho = 0$ | $n$ | WHSVM | $l_2$ SVM | $l_1$ SVM | frequency |
|---|---|---|---|---|---|
| | 50 | 0.186 (0.003) | 0.279 (0.003) | 0.206 (0.003) | 11/100 |
| | 100 | 0.154 (0.001) | 0.226 (0.002) | 0.169 (0.002) | 14/100 |
| | 200 | 0.145 (0.001) | 0.196 (0.001) | 0.149 (0.001) | 17/100 |
| | Bayes | | 0.133 | | |
| $\rho = 0.5$ | $n$ | WHSVM | $l_2$ SVM | $l_1$ SVM | frequency |
| | 50 | 0.173 (0.003) | 0.248 (0.003) | 0.190 (0.003) | 12/100 |
| | 100 | 0.159 (0.002) | 0.216 (0.002) | 0.167 (0.002) | 16/100 |
| | 200 | 0.143 (0.001) | 0.188 (0.001) | 0.147 (0.001) | 20/100 |
| | Bayes | | 0.130 | | |

TABLE 2
*Simulation example 2: The true model obeys the weak heredity principle. Compare the classification accuracy of the WHSVM, $l_2$ SVM and $l_1$ SVM. The numbers in parentheses are standard errors and the frequency is the number of times the fitted $l_1$ SVM obeys the weak heredity principle in 100 replications.*

| $\rho = 0$ | $n$ | WHSVM | $l_2$ SVM | $l_1$ SVM | frequency |
|---|---|---|---|---|---|
| | 50 | 0.248 (0.003) | 0.303 (0.003) | 0.273 (0.003) | 11/100 |
| | 100 | 0.198 (0.002) | 0.253 (0.002) | 0.216 (0.002) | 19/100 |
| | 200 | 0.163 (0.001) | 0.215 (0.002) | 0.183 (0.001) | 22/100 |
| | Bayes | | 0.142 | | |
| $\rho = 0.5$ | $n$ | WHSVM | $l_2$ SVM | $l_1$ SVM | frequency |
| | 50 | 0.199 (0.001) | 0.242 (0.002) | 0.220 (0.002) | 11/100 |
| | 100 | 0.164 (0.001) | 0.211 (0.001) | 0.181 (0.001) | 14/100 |
| | 200 | 0.143 (0.001) | 0.184 (0.001) | 0.154 (0.001) | 23/100 |
| | Bayes | | 0.121 | | |

Table 2 summarizes the simulation results. The WHSVM performs significantly better than the $l_1$ SVM and the $l_2$ SVM. The last column in Table 2 shows the frequency that the fitted $l_1$ SVM obeys the weak heredity principle. Again, these frequencies are pretty low.

**Simulation example 3.** Examples 1 and 2 have demonstrated the benefits of recognizing the effect heredity. It would be interesting to investigate the performance of the SHSVM and the WHSVM when the true model actually violates the heredity principle. To this end, we considered the third example. We generated 5 explanatory variables and simulated the class labels from

$$\log \left( \frac{\Pr(y = 1 | z_1, \ldots, z_5)}{\Pr(y = -1 | z_1, \ldots, z_5)} \right) = 3z_1 + 2.5z_2 + 2z_3 z_4 + 1.5 z_4 z_5 + 1.$$

We fitted the SHSVM, WHSVM, $l_2$ SVM and $l_1$ SVM using the predictor set $\{z_j, z_r z_j, z_j^2\}$, $r, j = 1, \ldots, 5$. Since the true model is sparse, we expect that the $l_2$ SVM has the worst performance. This is confirmed by the simulation results in Table 3. We see that there is basically no difference between the WHSVM and the $l_1$ SVM. This observation suggests that it does not hurt to enforce the heredity principle along with the sparsity, even when the true model does not obey the heredity principle.

TABLE 3
*Simulation example 3: Compare the SHSVM, WHSVM, $l_2$ SVM and $l_1$ SVM when the true model obeys no heredity principle. The numbers in parentheses are standard errors.*

| $\rho = 0$ | $n$ | SHSVM | WHSVM | $l_2$ SVM | $l_1$ SVM |
|---|---|---|---|---|---|
|  | 50 | 0.171 (0.002) | 0.164 (0.002) | 0.203 (0.003) | 0.172 (0.003) |
|  | 100 | 0.147 (0.002) | 0.140 (0.002) | 0.173 (0.002) | 0.143 (0.002) |
|  | 200 | 0.131 (0.001) | 0.125 (0.001) | 0.151 (0.001) | 0.127 (0.001) |
|  | Bayes | 0.113 | | | |
| $\rho = 0.5$ | $n$ | SHSVM | WHSVM | $l_2$ SVM | $l_1$ SVM |
|  | 50 | 0.138 (0.002) | 0.137 (0.002) | 0.156 (0.002) | 0.139 (0.002) |
|  | 100 | 0.119 (0.001) | 0.115 (0.001) | 0.134 (0.002) | 0.115 (0.001) |
|  | 200 | 0.109 (0.001) | 0.104 (0.001) | 0.128 (0.001) | 0.105 (0.001) |
|  | Bayes | 0.093 | | | |

**Birth weight data.** We test the proposed structured SVMs on the birth weight data that concern the birth weight of 189 infants at a US hospital [16]. The problem of interest is to predict if the birth weight is lower than 2.5 kg. There are 8 explanatory variables depending upon mother's age ($age$), weight ($lwt$), race ($race$), smoking status ($smoke$), number of previous premature labors ($ptl$), history of hypertension ($ht$), uterine irritability ($ui$), and number of physician visits in the first trimester ($ftv$). The variables $age$ and $lwt$ are continuous while dummy variables were used to represent the discrete-valued variables. Then the predictor set was generated as in the simulation models except that the quadratic effects of dummy variables were not included. For dummy variables, the heredity principles were applied to the group level. Because the sample size is only 189, we used 5-fold cross-validation to estimate the classification error of each method.

As can be seen from Table 4, the structured SVMs significantly outperforms both the $l_2$ SVM and the $l_1$ SVM. The best $l_1$ SVM model identifies 10 variables including $age^2$, $age \cdot lwt$, $age \cdot ftv$, $lwt^2$, $lwt \cdot race$, $lwt \cdot smoke$, $lwt \cdot ptl$, $lwt \cdot ht$, $lwt \cdot ui$, and $lwt \cdot ftv$. This model does not satisfy the heredity principles, because, for instance, $age^2$ and $age \cdot lwt$ are included without their parent factor $age$. The frequencies of the $l_1$ SVM model satisfying the strong and weak heredity principles were 10/20 and 14/20, respectively. The model selected by the WHSVM includes $age$, $lwt$, and $ftv$ together with the 10 variables in the $l_1$ SVM model. The SHSVM model includes additional variables $race$, $smoke$, $ptl$, $ht$, $ui$, and $ftv$ in the WHSVM model.

One might wonder which heredity SVM should be used in this real data example. If the modeler does not have a strong preference in using either strong or weak heredity principle, the data suggest that the WHSVM is perhaps better than the SHSVM, since they have very similar classification performance and the WHSVM uses less variables.

TABLE 4
*Birth weight data: average five-fold cross validation errors with standard errors (reported in parentheses) based on 30 replications.*

| SHSVM | WHSVM | $l_2$ SVM | $l_1$ SVM |
|---|---|---|---|
| 0.291 (0.002) | 0.294 (0.002) | 0.307 (0.002) | 0.305 (0.001) |

## 3. Nonparametric Heredity SVMs

In the previous section we have discussed the heredity principle when each effect is represented by a single predictor. In many real world applications, we often need to nonparametrically model the main and interaction effects. Let us consider the following model where the class label $y$ and explanatory variables $z_1, z_2, \ldots, z_q$ are related through

$$\log\left(\frac{\Pr(y=1|z_1,\ldots,z_q)}{\Pr(y=-1|z_1,\ldots,z_q)}\right) = \sum_{j=1}^{q} f_j(z_j) + \sum_{r,j=1}^{q} f_{rj}(z_r, z_j). \qquad (3.1)$$

We have omitted the constant term for simplicity. The main effect of variable $z_j$ is $f_j(z_j)$ and the interaction effect between variables $z_r$ and $z_j$ is $f_{rj}(z_r, z_j)$. Obviously, the above model is a generalization of the popular *Generalized Additive Model* [12]. The model (3.1) can be more appropriate than the generalized additive model if interaction effects cannot be ignored.

Under the strong heredity, for the interaction effect $f_{rj}(z_r, z_j)$ to be active both its parent effects, $f_r(z_r)$ and $f_j(z_j)$, should be active, whereas under the weak heredity only one of its parent effects needs to be active. In this section we develop a method that can automatically identify significant effects while respecting the heredity principle.

### *3.1. Imposing heredity principles*

If we assume $f_j(z_j) = \beta_j z_j$ and $f_{rj}(z_r, z_j) = \beta_{rj} z_r z_j$, then the model reduces to the parametric case. We show here that the parametric assumption is not necessary in order to implement the heredity principle by using the SVS framework. Suppose that we have found a good initial estimate of the full model (3.1) and we denote the initial estimates by $\hat{f}_j(z_j)$ and $\hat{f}_{rj}(z_r, z_j)$. We assign scaling parameters $\theta_j$ to $f_j(z_j)$ and $\theta_{rj}$ to $f_{rj}(z_r, z_j)$. The SHSVM can be formulated as follows

$$\min \sum_{i=1}^{n} \left[1 - y_i\left(\sum_{j=1}^{q} \hat{f}_j(z_j)\theta_j + \sum_{r,j=1}^{q} \hat{f}_{rj}(z_r, z_j)\theta_{rj}\right)\right]_+ \quad (3.2)$$

subject to
$$\sum_{j=1}^{q} \theta_j + \sum_{r,j=1}^{q} \theta_{rj} \leq M$$
$$\theta_j \geq 0 \quad \theta_{rj} \geq 0 \quad \forall r, j$$
$$\theta_{rj} \leq \theta_r \quad \text{and} \quad \theta_{rj} \leq \theta_j \quad \forall r, j. \qquad (3.3)$$

Likewise, we define the WHSVM as

$$\min \sum_{i=1}^{n} \left[ 1 - y_i(\sum_{j=1}^{q} \hat{f}_j(z_j)\theta_j + \sum_{r,j=1}^{q} \hat{f}_{rj}(z_r, z_j)\theta_{rj}) \right]_+ \quad (3.4)$$

subject to
$$\sum_{j=1}^{q} \theta_j + \sum_{r,j=1}^{q} \theta_{rj} \leq M$$
$$\theta_j \geq 0 \quad \theta_{rj} \geq 0 \quad \forall r, j$$
$$\theta_{rj} \leq \theta_r + \theta_j \quad \forall r, j. \quad (3.5)$$

The final classifier is $\text{Sign}\left(\hat{f}_j(z_j)\hat{\theta}_j + \sum_{r,j=1}^{q} \hat{f}_{rj}(z_r, z_j)\hat{\theta}_{rj}\right)$. The linear inequalities in (3.3) guarantee that the SHSVM obeys the strong heredity principle. Similarly, the linear inequalities in (3.5) guarantee that the WHSVM obeys the weak heredity principle. Moreover, solving the scaling parameters is a linear program.

### 3.2. Computing the initial estimator

There are many nonparametric estimation methods that can give us a good initial estimator of the model (3.1). The choice of the estimation method is not essential for using the SHSVM and the WHSVM. In this work, for computational considerations, we obtain the initial estimates by using penalized B-splines [6]. Penalized B-splines have been widely used in statistics for nonparametric function estimation (cf. [8], [11] and [17]). For each variable $z_j$, we take a basis of B-spline functions $b_{j,k}(z_j)$ for $k = 1, 2, \ldots, N_j$ for representing the function $f_j(z_j)$. Then the $N_r \times N_j$ dimensional *tensor product basis* defined by

$$g_{k_1, k_2}(z_r, z_j) = b_{r,k_1}(z_r)b_{j,k_2}(z_j), k_1 = 1, 2, \ldots, N_r \quad \text{and} \quad k_2 = 1, 2, \ldots, N_j$$

can be used for representing the interaction effect $f_{rj}(z_r, z_j)$. With B-spline basis functions at hand, we can compute the $l_2$ SVM estimate of the model (3.1) by minimizing

$$\sum_{i=1}^{n} \left[ 1 - y_i \left( \alpha_0 + \sum_{j=1}^{q} \sum_{k=1}^{N_j} \alpha_{jk} b_{j,k}(z_j) \right. \right.$$
$$\left. \left. + \sum_{r,j=1}^{q} \sum_{k_1=1}^{N_r} \sum_{k_2=1}^{N_j} \alpha_{rjk_1k_2} b_{r,k_1}(z_r) b_{j,k_2}(z_j) \right) \right]_+ + \lambda \|\alpha\|_2^2,$$

where $\|\alpha\|_2^2 = \sum_{k=1}^{N_j} \alpha_{jk}^2 + \sum_{r,j=1}^{q} \sum_{k_1=1}^{N_r} \sum_{k_2=1}^{N_j} \alpha_{rjk_1k_2}^2$. Then the initial estimates are

$$\hat{f}_j(z_j) = \sum_{k=1}^{N_j} \hat{\alpha}_{jk} b_{j,k}(z_j),$$

$$\hat{f}_{rj}(z_r, z_j) = \sum_{k_1=1}^{N_r} \sum_{k_2=1}^{N_j} \hat{\alpha}_{rjk_1k_2} b_{r,k_1}(z_r) b_{j,k_2}(z_j).$$

In computing the initial estimates, although there are $q$ variables, the actual dimension of the predictor set is $\sum_{j=1}^{q} N_j + \sum_{r,j=1}^{q} N_j N_r$, which could be a large number. The quadratic penalty not only regularizes the nonparametric fit but also allows for an efficient implementation through singular value decomposition [10]. Let $\mathcal{B}$ denote the basis functions in the predictor set. We need to solve

$$(\hat{\alpha}_0, \hat{\alpha}) = \arg\min_{\alpha_0, \alpha} \sum_{i=1}^{n} [1 - y_i(\alpha_0 + \mathcal{B}_i \alpha)]_+ + \lambda \alpha^T \alpha.$$

Suppose the singular value decomposition of $\mathcal{B}$ is $\mathcal{B} = UDV^T = RV^T$ where $R$ is a $n \times n$ matrix, then we solve

$$(\hat{\gamma}_0, \hat{\gamma}) = \arg\min_{\gamma_0, \gamma} \sum_{i=1}^{n} [1 - y_i(\alpha_0 + R_i \gamma)]_+ + \lambda \gamma^T \gamma,$$

and $\hat{\alpha} = V\hat{\gamma}$ and $\hat{\alpha}_0 = \hat{\gamma}_0$. See theorem 1 in [10]. Therefore, the computations can be done in a $n$ dimensional space instead of the original high-dimensional predictor space.

### 3.3. Numerical examples

We now present some numerical examples to demonstrate the performance of the nonparametric heredity SVMs. We compared the nonparametric heredity SVMs with the $l_2$ SVM and the Gaussian kernel SVM. In all examples, the $l_2$ SVM was fitted using the same B-Spline basis functions for fitting the nonparametric heredity SVMs.

**Simulation example 4.** We first generated explanatory variables $z_1,\ldots,z_5$ from a multivariate normal distribution in which the correlation between $z_r$ and $z_j$ is $0.5^{|r-j|}$. We considered a sparse model where the class labels were generated from a logistic regression model

$$\log\left(\frac{\Pr(y=1|z_1,\ldots,z_5)}{\Pr(y=-1|z_1,\ldots,z_5)}\right) = f_1(z_1) + f_2(z_2) + f_{12}(z_1, z_2) + 1.$$

The true model obeys the strong heredity principle. We used five B-spline basis functions $\{b_{j,1}(z_j),\ldots,b_{j,5}(z_j)\}$ to represent each $f_j(z_j)$, and the interaction effect $f_{rj}(z_r, z_j)$ was represented by the tensor product basis functions $\{b_{r,1}(z_r)b_{j,1}(z_j),\ b_{r,1}(z_r)b_{j,2}(z_j),\ \ldots,\ b_{r,5}(z_r)b_{j,5}(z_j)\}$. The representing coefficients ($\alpha$) were chosen as follows: (i) Coefficients of the 5 basis functions for $f_1(z_1)$ are $(2.1, -2.9, 0.3, 2.7, -0.1)$, (ii) coefficients of the 5 basis functions for $f_2(z_2)$ are $(-2.8, -1.2, 1.8, 1.7, -0.8)$, and (iii) coefficients of the 25 basis functions for $f_{12}(z_1, z_2)$ are $(-2.4, -0.1, 0.6, 3, 2.8, -0.9, 0.3, 1, -0.9, -1.3, 0.9, 2.3, 1.9, 0.8, -0.2, 1.2, 2.1, 1.0, -0.8, -1.7, -0.8, -1.2, 2.1, -2.8, 0.1)$.

It should be mentioned that in this model the dimension of the predictor set is 275. On the other hand, this model is very sparse in terms of the number of active effects (only three active effects). We simulated a training sample of size 100 from the above model and collected an independent test sample of size 10000 to compute the generalization error of each competitor. The simulation was repeated 100 times. The simulations results are summarized in Table 5 from which two interesting observations can be made. First we see that the $l_2$ SVM actually does better than the Gaussian kernel SVM in this example. This observation suggests that although the Gaussian kernel SVM is perhaps the most popular nonparametric SVM classifier, it is not always the best choice in all problems. Second and more importantly, the SHSVM is clearly the winner among all three competitors.

TABLE 5
*Compare the SHSVM, the $l_2$ SVM and the Gaussian kernel SVM when the true model obeys the strong heredity principle. The numbers in parentheses are standard errors.*

| SHSVM | $l_2$ SVM | GK-SVM | Bayes |
|---|---|---|---|
| 0.209 (0.001) | 0.214 (0.001) | 0.218 (0.001) | 0.202 |

TABLE 6
*Compare the WHSVM, the $l_2$ SVM and the Gaussian kernel SVM when the true model obeys the strong heredity principle. The numbers in parentheses are standard errors.*

| WHSVM | $l_2$ SVM | GK-SVM | Bayes |
|---|---|---|---|
| 0.217 (0.001) | 0.226 (0.001) | 0.222 (0.001) | 0.197 |

**Simulation example 5.** In this example we considered the same setup in example 4, except that the class labels were generated from a logistic regression model

$$\log \left( \frac{\Pr(y = 1|z_1, \ldots, z_5)}{\Pr(y = -1|z_1, \ldots, z_5)} \right) = f_1(z_1) + f_2(z_2) + f_{15}(z_1, z_5) + f_{23}(z_2, z_3) - 1.$$

Hence this model obeys the weak heredity principle. As in example 4, we used B-splines to model each effect. The representing coefficients are chosen as follows: (i) Coefficients of the 5 basis functions for $f_1(z_1)$ are $(3.0, -2.5, 2.0, -1.5, 1.0)$, (ii) coefficients of the 5 basis functions for $f_2(z_2)$ are $(1.5, 2.0, -3.0, -2.5, -2.0)$, (iii) coefficients of the 25 basis functions for $f_{15}(z_1, z_5)$ are $(7.1, -9.8, 1.1, 9.0, -0.3, -8.1, -0.4, 2.0, 10, 9.4, -3.1, 1.0, 3.2, -3.1, -4.3, 3.1, 7.7, 6.2, 2.7, -0.7, 3.9, 6.8, 3.4, -2.5, -5.6)$, and (iv) coefficients of the 25 basis functions for $f_{23}(z_2, z_3)$ are $(-2.6, -3.8, 7.0, -9.4, 0.5, -9.2, -4.0, 6.1, 5.6, -2.7, 5.5, 9.3, -5.4, 9.1, -2.8, 5.1, 3.9, 6.6, -0.6, 6.8, 0.8, 8, -3.6, -2.5, -6)$.

As can be seen from Table 6, in this example the Gaussian kernel SVM outperforms the $l_2$ SVM, but the best performance is given by the WHSVM.

**South African Heart Disease Data.** Here we demonstrate the utility of the nonparametric heredity SVMs through an analysis of the South African heart disease data [11] which consist of 462 samples of 9 risk factors (8 continuous and 1 binary). The responses indicates the presence of heart disease. Previous studies of this data suggest that nonparametric functions should be used to model the effects of these 9 risk factors. We first used the popular Gaussian kernel SVM to analyze the data whose classification error can be used as a good benchmark for comparison. To fit the SHSVM and the WHSVM, we used B-splines to flexibly model the main effects of 8 continuous risk factors and use the tensor product basis functions of B-splines to model the interaction effects. In total, there are 33 basis functions and 480 basis functions used for representing the main effects and the interaction effects, respectively. Since we did not have an independent test set, we found the smallest 5-fold cross-validation error of each competitor. Then we repeated the whole procedure 30 times and reported the average 5-fold cross-validation errors. As can be seen from Table 7, the SHSVM does significantly better than the Gaussian kernel SVM.

Table 7
South African heart disease data: average 5-fold cross-validation errors based on 30 replications. The numbers in parentheses are standard errors.

| SHSVM | WHSVM | GK-SVM |
|---|---|---|
| 0.267 (0.001) | 0.270 (0.001) | 0.276 (0.001) |

## 4. Discussion

In this paper we have developed a unified framework for simultaneously incorporating the heredity principle and sparsity into the support vector machine. By adopting the scaling parameter idea from the nonnegative garrote, we have shown that both strong and weak heredity principles can be enforced by a set of linear inequality constraints on the scaling parameters. Our approach is computationally efficient, as the optimization problem a linear program. Moreover, we have also extended the framework to handle nonparametric models, which shows the flexibility of our method. The encouraging numerical results suggest that the newly proposed method is a useful addition to the classification toolbox.

To fix the main idea, we have used the penalized $l_2$ SVM to construct the initial classifier. Based on our experience, this choice of initial classifier worked quite well even when the dimension of predictors exceeds the sample size. It is possible to further improve the heredity SVMs by using better initial classifiers in certain problems.

Finally, we comment on the path-based computation of the structured SVMs. Yuan and Lin [22] showed that the solution path of the original nonnegative garrote is piecewise linear and constructed an efficient algorithm for building its whole solution path. One may expect the same is true for the garrote SVM. With the heredity constraints, the solution paths of $\theta$s will remain piece-wise linear as a function of their $l_1$ norm. However, the path-following algorithm will become considerably more complicated. It is not clear if computing the whole solution path will provide us considerable computational savings, compared with running linear programming for a grid of tuning parameters.